# AN ENERGY-BASED COMPARATIVE ANALYSIS OF COMMON APPROACHES TO TEXT CLASSIFICATION IN THE LEGAL DOMAIN

Sinan Gultekin, Achille Globo, Andrea Zugarini, Marco Ernandes, and Leonardo Rigutini

Department of Hybrid Linguistic Technologies expert.ai, Siena, Italy

## ABSTRACT

*Most Machine Learning research evaluates the best solutions in terms of performance. However, in the race for the best performing model, many important aspects are often overlooked when, on the contrary, they should be carefully considered. In fact, sometimes the gaps in performance between different approaches are neglectable, whereas factors such as production costs, energy consumption, and carbon footprint must take into consideration. Large Language Models (LLMs) are extensively adopted to address NLP problems in academia and industry. In this work, we present a detailed quantitative comparison of LLM and traditional approaches (e.g. SVM) on the LexGLUE benchmark, which takes into account both performance (standard indices) and alternative metrics such as timing, power consumption and cost, in a word: the carbon-footprint. In our analysis, we considered the prototyping phase (model selection by training-validation-test iterations) and in-production phases separately, since they follow different implementation procedures and also require different resources. The results indicate that very often, the simplest algorithms achieve performance very close to that of large LLMs but with very low power consumption and lower resource demands. The results obtained could suggest companies to include additional evaluations in the choice of Machine Learning (ML) solutions.*

## 1. INTRODUCTION

Over the past decade, we have observed a critical paradigm shift in the field of NLP. The increasing diffusion of end-to-end approaches led to the development of a broad set of Large Language Models (LLMs) based on different neural network architectures and consisting of billions of parameters. Given their huge training and deployment costs, these giant models are typically exclusive to the handful of global companies (i.e., Google, Microsoft) that can sustain such costs. They are typically released as pre-trained models and require a fine-tuning step to refine the model based on the customer's requirements. However, they require vast amounts of resources to operate in terms of hardware and energy. Most academics, data scientists, or insiders often ignore aspects of energy consumption, but the increasing energy-hungry computation trend raises some relevant concerns. From an ethical and social point of view, we are all witnesses to severe climate change due to pollution and $CO_2$ emissions. From an economic and industrial point of view, however, in recent years, the energy cost has reached extremely high levels, and having good light Machine Learning solutions can be of vital for companies.

In this article we present a comparative analysis of two widely used families of text classification models in terms of performance and power consumption. In particular, the investigation aims to explore the balance between the performance and carbon footprint of several models based on (1) Large Language Models (LLM) and on (2) Support Vector Machines (SVM) when employed in a vertical domain. On the performance side, the standard classification metric F1 is considered, while on the green side, the energy consumption (KWh), the estimated costs (€) and $CO_2$ production are valued. The tests were carried out using the LexGLUE benchmark and the results show that, in many cases, lightweight models obtain excellent performance at significantly lower costs. These results suggest further in-depth studies on the use of Deep Learning approaches in industry and underline the need to consider several aspects in addition to the quality of the predictions when selecting the best ML solution in NLP projects.

The paper is organized as follows. Section 2 reports the related works and provides some of the reasons that led us to carry out this analysis and experimentation. In Section 3, the details of the investigation are described, such as the models and the datasets employed, while in Section 4, we report the results of the experiments and outline the emerging considerations. Finally, Section 5 draws conclusions and possible ideas for future works.

## 2. RELATED WORK AND MOTIVATION

Training and deployment cost of deep neural networks have escalated enormously in the last decade, which drives modern ML models into the energy-hungry trail. These developments lead some researchers to draw attention to models' efficiency and potential adaptations. Many studies have addressed the problem of model size compression through different approaches, such as knowledge distillation [15], pruning [24], quantization [9], and vocabulary transfer [8]. However, although in many areas, a communication strategy based on the green-friendly is increasingly present (as in Google [1]and Amazon [2]), in the Artificial Intelligence (AI) research field, this topic has not yet played an important role, and it is not receiving the proper attention. In recent years the topic of the eco-sustainability of artificial intelligence started to be addressed but despite the attempt to highlight the importance of environmental consideration, only a few works appear in the literature [14]. In [21], the authors report a comparison between some models of neural networks used in NLP in terms of energy consumption and $CO_2$ production. In the paper, they describe how the energy required for one training cycle of a transformer-based NLP model produces much more $CO_2$ than the average human produces in a year. In this work, however, the analysis does not include lightweight methods (such as SVM), and it does not correlate the costs with the performance. In [17], the authors propose a deep meditation for the eco-sustainability of AI and outline an essential prevalence of Red AI compared to Green AI in the scientific field. In particular, they analyze a sample of papers published in top AI conferences and show how the efficiency topic is highly uncommon. The results are sketched in the figure 1.

At the same time, some tools for evaluating the Carbon-Footprint have been presented

---

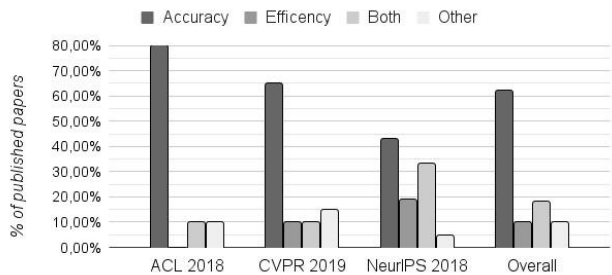

Fig.1: Trend of accuracy and efficiency in AI papers. The charts were recreated with data from [17].

in the literature, as by [12] and [11]. In particular, in [11], the "codecarbon" library is presented, and nowadays, it is one of the most used tools for measuring an algorithm's energy consumption and carbon footprint.

We were prompted to our investigation by the fact that, from the analysis of the works in the literature addressing the eco-sustainability of AI, none of them was proposing a combined analysis related to performance with energy consumption, costs, and carbon footprint in a real business scenario. Instead, we believe that an analysis of this type should always be carried out in the evaluation of Artificial Intelligence solutions since this trend towards ever larger models (and therefore an even more energy-hungry computation) raises significant concerns and, in many real cases, is not necessary. From an ethical and social point of view, we can all see the serious results of climate change due to pollution and, in particular, $CO_2$ emissions. Most countries are developing alternative solutions to fossil fuels, but a more conscious use of resources is also essential. The world of AI-related companies must also do its part and focus on technologies and solutions that are also environmentally friendly without, of course, losing in performance. Besides that, a problem of democratic access to resources exists. The race to increase the size of neural network models has resulted in them being the prerogative of a few global IT companies, leaving out most of both university and private research laboratories and small companies. This cycle is also known as "the rich get richer". On the other hand, from an economic and industrial point of view, it is known that the energy cost has reached extremely high in recent years. For this reason, finding lightweight AI solutions can represent significant cost savings, which can be crucial for the survival of companies.

For all these reasons, we believe that the presented analysis can be instrumental in suggesting that, in many real cases, even other aspects can be considered in addition to performance when choosing an AI solution.

## 3. THE INVESTIGATION

In this article we present a comparative analysis of two widely used families of text classification models in terms of performance and power consumption. The investigation aims to explore the balance between the performance and carbon footprint of different models based on (1) Large Language Models (LLM) and on (2) Support Vector Machines (SVM) when employed in a vertical domain. Our goal is to reproduce a typical situation in the real world where usually the analyzed documents concern a specific domain of interest (e.g., financial, legal, health). In particular, in this study, we chose to address the "legal" area, selecting a standard benchmark for this sector, the LexGLUE.

## 3.1. The Benchmark

Following the spread of multitask benchmarks in the NLP field, such as GLUE and Super-GLUE, the LexGLUE Benchmark [5] was recently released. The LexGLUE (Legal General Language Understanding Evaluation) benchmark is a collection of seven datasets focused on the legal domain and built for evaluating model performance across a diverse set of legal NLP tasks. The first version of the benchmark[2]only covers the English language. However, more datasets, tasks, and languages are expected to be added in later versions of LexGLUE as new legal NLP datasets become available. The seven datasets were built using different legal sources, including the European Court of Human Rights (ECtHR), the U.S. Supreme Court (SCOTUS), the European Union legislation (EUR-LEX), the U.S. Security Exchange Commission (LEDGAR), the Terms of Service from famous online platforms (Unfair-ToS) and a Case Holdings on Legal Decisions (CaseHOLD). The dataset details are summarized in Table 1 and they are deeply described in the original paper [5].

Table 1: Statistics about the seven datasets included in the LexGLUE benchmark.

| Dataset | Data Type | Task | Train/Validation/Test | Classes |
|---|---|---|---|---|
| ECtHR (Task A) | ECHR | Multi-label classification | 9,000/1,000/1,000 | 10+1 |
| ECtHR (Task B) | ECHR | Multi-label classification | 9,000/1,000/1,000 | 10+1 |
| SCOTUS | US Law | Multi-class classification | 5,000/1,400/1,400 | 14 |
| EUR-LEX | EU Law | Multi-label classification | 55,000/5,000/5,000 | 100 |
| LEDGAR | Contracts | Multi-class classification | 60,000/10,000/10,000 | 100 |
| Unfair ToS | Contracts | Multi-label classification | 5,532/2,275/1,607 | 8+1 |
| CaseHOLD | US Law | Multiple choice QA | 45,000/3,900/3,900 | n/a |

## 3.2. Models

For the study, we choose two families of models largely-used in the text classification task: LLM and SVM. From the first group (LLM), we selected three BERT-based models: BERT, LegalBERT and DistilBERT; from the second group (SVM), we chose two feature representations: the classic Bag-Of-Word (BoW) and an advanced representation enriched with linguistic and semantic features. In the experiments, where possible, we reproduced the same configurations reported in the original LexGLUE experimentation [5].

BERT-based models – BERT [7] is one of the most popular LLMs and it is based on the transformer architecture. It is available as a model pre-trained on a massive dataset of general-purpose documents, thus representing a good generic language model. It has reported excellent results in the field of text analysis and NLP but, being based on a large deep neural network, requires a lot of resources to run.

Moreover, when dealing with a specific domain, having a language model that builds language statistics from the terminology used in the particular domain could be helpful. Thus, some variants of BERT have been proposed in the literature where they have been re-trained on domain-specific documents. Since, we faced the legal domain, we included also LegalBERT [4] in the comparative analysis, a derived BERT model which has been pre-trained on legal corpora such as legislation, contracts, and court cases.

---

[2] https://github.com/coastalcph/lex-glue

Finally, since our analysis addresses energy consumption and this is closely related to the size of the model, we also included DistilBERT [16] in the evaluation, a scaled-down version of the original BERT model obtained by using distillation.

SVM-based approaches – Support Vector Machines (SVMs) [6] are well-established Machine Learning models, that have been widely used also in text categorization for decades [13,19]. They work by identifying a small optimal subset of the training examples that best define the separation hyperplane. Furthermore, they involve the use of kernels that allow the identification of nonlinear hyperplanes of separations.

As a first SVM-based approach, we selected a very simple and basic setup consisting of a linear kernel SVM on top of a Bag-Of-Word (BoW) representation which has been the most widely used approach for text categorization problems for many years [13]. Furthermore, we also considered an approach that combines the standard text representation (BoW) with additional linguistic and semantic features. This approach has also been widely used in past years, demonstrating good results in text classification problems [1,18,23]. We also selected such an approach to test whether the inclusion of external linguistic knowledge in the feature space can lead to a reduction in model complexity (and therefore a reduction in energy consumption) without a significant performance loss. In this approach, a preliminary NLP step produces a set of linguistic and semantic features (e.g. lemmas, Part-Of-Speech tags, concepts, etc.) that is combined with the standard Bag-Of-Word representation. The new augmented feature space is then used to train Machine Learning models. For the NLP analysis, we used the expert.ai hybrid natural language platform, while a linear SVM was used as the on-top ML classifier. The expert.ai natural language platform consists in an integrated environment for deep language understanding and provides a complete natural language workflow with end-to-end support for annotation, labeling, model training, testing and workflow orchestration [3].

In the paper we will refer to these two approaches as $SVM_{bow}$ and $SVM_{nlp}$, respectively.

### 3.3. Experimental Setup

The comparative analysis was carried out using both performance-oriented metrics and indices related to the eco-friendly. For the performance, we used the standard F1 score (both micro mF1 and macro MF1), while for the eco-friendly we estimated the energy consumption (KWh), the costs (€) and the carbon footprint ($CO_2$) consumed by each approach. Furthermore, for a better understanding of the costs/benefits of the analyzed approaches, we have also separately evaluated the cost in energy terms of the prediction phase alone. In fact, in the industrial field, this step is the one that is performed with much higher frequency than the training phase. For the evaluation of the energy consumption, we employed "codecarbon"[5], a widely used library that allows measuring the energy consumed by the system in executing a sequence of instructions, also including the possible use of GPU [11]. In particular, we replicated the same experiments reported in the LexGLUE article by including the "codecarbon" library instructions for measuring energy-related indices directly in the authors' code. For the $SVM_{nlp}$ approach, we also considered the step of the NLP analysis. Finally, we did not include the CaseHOLD dataset since, unlike the others, it was released for a Question Answering (QA) task that significantly differs from text classification. Experiments were carried out on an Intel Xeon processor-based server

---

[3] https://www.expert.ai/products/expert-ai-platform/

[5] https://github.com/mlco2/codecarbon/

with 503GB of RAM equipped with an NVIDIA RTX A6000 GPU with 49GB of dedicated RAM.

## 4. EXPERIMENTAL RESULTS

Typically, NLP project development involves two main phases: (a) model training and evaluation, where data scientists iteratively perform training-validation-test steps to assess the solution (R&D phase) and (b) final delivery and production, in which the selected model is released and used in a production environment. Therefore, we performed two types of investigations. First, we compared models in terms of performance and energy consumption during a typical train/validation/test procedure. In the second, we compared the energy and time required by the models to make predictions on a fixed number of documents.

### 4.1. R&D Scenario

In the first analysis, we simulated the R&D phase of a project. This activity is involved in the system's initial setup and is often repeated several times. The number of trials depends on the project's characteristics and nuances; in many cases, it can be significant, making the estimate of the effort unreliable. In the following, we report the comparative analysis individually for each dataset in terms of (a) performance, using the F1 score, both micro (mF1) and macro (MF1) averaging, and (b) energy consumption (KWh), costs (€) and carbon footprint ($CO_2$) estimated for each experiment.

ECtHR Datasets – The results of the tests on the two European Court of the Human Rights (ECtHR) datasets [2] are reported in Table 2. In both datasets, the $SVM_{nlp}$ approach results to be the most eco-friendly while remaining with the same performance as $SVM_{bow}$. In particular, both SVM-based models show quite lower performance than BERT and LegalBERT but the latter report at least 40 and up to 75 times higher energy consumption of the $SVM_{nlp}$ approach. On the other hand, DistilBERT reports intermediate consumption (from 3 to 20 times higher than $SVM_{nlp}$) but with performances in some cases even lower.

Table 2: Classification performances and the energy consumption results of different models on ECtHR datasets.

|        | Metric | $SVM_{bow}$ | $SVM_{nlp}$ | BERT | LegalBERT | DistilBERT |
|--------|--------|-------------|-------------|------|-----------|------------|
| ECtHR A | mF1    | 0.65 | 0.65 | **0.71** | 0.70 | 0.62 |
|        | MF1    | 0.52 | 0.52 | **0.64** | **0.64** | 0.56 |
|        | KWh    | ×1.95 | 1 | ×73.93 | ×74.25 | ×23.98 |
|        | €      | ×1.95 | 1 | ×73.93 | ×74.25 | ×23.98 |
|        | $CO_2$ | ×1.32 | 1 | ×23.42 | ×23.52 | ×7.60 |
| ECtHR B | mF1    | 0.75 | 0.75 | **0.80** | **0.80** | 0.71 |
|        | MF1    | 0.65 | 0.65 | **0.73** | **0.73** | 0.61 |
|        | KWh    | ×1.56 | 1 | ×62.49 | ×36.56 | ×3.39 |
|        | €      | ×1.56 | 1 | ×62.49 | ×36.56 | ×3.39 |
|        | $CO_2$ | ×1.16 | 1 | ×21.83 | ×5.00 | ×1.78 |

EUR-LEX – Table 3 reports the assessment results of the European Union Legislation (EUR-LEX) dataset [3]. Even in this dataset, the $SVM_{nlp}$ model remains the most ecofriendly maintaining extremely acceptable performances. In particular, it demonstrates good performance with approximately half the power consumption of $SVM_{bow}$ and approximately three times lower than BERT-based approaches. In this case, however, the energy saving and pollution rates are proportionally lower than in the previous case.

LEDGAR – In Table 4, we report the results on the Labeled Electronic Data Gathering, Analysis, and Retrieval system (LEDGAR) dataset [22]. In this case, the $SVM_{nlp}$ approach reports both the best performance and the best power consumption values. In fact, it shows energy saving rates of up to 80 times compared to fully BERT-based approaches. DistilBERT also shows acceptable performance but with energy consumption still significantly higher than $SVM_{nlp}$.

Table 3: The classification performances and the energy consumption results of different models on EUR-LEX dataset.

|  | Metric | $SVM_{bow}$ | $SVM_{nlp}$ | BERT | LegalBERT | Distil-BERT |
|---|---|---|---|---|---|---|
| **EUR-LEX** | mF1 | 0.71 | 0.73 | 0.71 | 0.72 | 0.74 |
| | MF1 | 0.51 | 0.50 | 0.57 | 0.57 | 0.46 |
| | KWh | ×1.85 | 1 | ×4.81 | ×4.89 | ×1.91 |
| | € | ×1.85 | 1 | ×4.81 | ×4.89 | ×1.91 |
| | $CO_2$ | ×1.12 | 1 | ×1.56 | ×1.58 | ×1.62 |

Table 4: Classification performance and the energy consumption results of different models on LEDGAR dataset.

|  | Metric | $SVM_{bow}$ | $SVM_{nlp}$ | BERT | LegalBERT | Distil-BERT |
|---|---|---|---|---|---|---|
| **LEDGAR** | mF1 | 0.88 | 0.89 | 0.88 | 0.88 | 0.88 |
| | MF1 | 0.82 | 0.84 | 0.82 | 0.83 | 0.81 |
| | KWh | ×1.67 | 1 | ×53.21 | ×77.71 | ×24.28 |
| | € | ×1.67 | 1 | ×53.21 | ×77.71 | ×24.28 |
| | $CO_2$ | ×1.34 | 1 | ×20.05 | ×29.28 | ×9.15 |

SCOTUS – The results obtained on the US Supreme Court (SCOTUS) dataset [20] are reported in Table 5. They confirm the same trend as the previous case. Furthermore, in this case, the $SVM_{nlp}$ approach significantly outperforms the other models while remaining the best option in terms of consumption. In particular, it shows F1 values about 10 points higher than both BERT and DistilBERT, and 3 points higher than LegalBERT. These results are obtained while maintaining energy savings of approximately 2 times compared to DistilBERT and 15-20 times compared to LegalBERT and BERT, respectively.

Table 5: Classification performances and the energy consumption results of different models on SCOTUS dataset.

|  | Metric | $SVM_{bow}$ | $SVM_{nlp}$ | BERT | LegalBERT | Distil-BERT |
|---|---|---|---|---|---|---|
| **SCOTUS** | mF1 | 0.78 | 0.79 | 0.68 | 0.76 | 0.68 |
| | MF1 | 0.69 | 0.70 | 0.58 | 0.67 | 0.57 |
| | KWh | ×1.33 | 1 | ×19.36 | ×15.10 | ×1.95 |
| | € | ×1.33 | 1 | ×19.36 | ×15.10 | ×1.95 |
| | $CO_2$ | ×1.08 | 1 | ×5.28 | ×4.12 | ×1.53 |

Unfair ToS – The results of the tests on the Unfair Terms of Services (Unfair ToS) dataset [10] can be seen in Table 6. Unfair-ToS is the smallest data set in the LexGLUE benchmark. The tests show that the $SVM_{bow}$ model reports the best values regarding energy savings while maintaining performance very close to the best models. However, even in this case, the $SVM_{nlp}$ model stands up to the excellent competition in performance and energy savings with very close outcomes.

Although BERT-based models report the best performances, a serious concern may arise about their energy consumption which is an average of 30 times and 60 times compared to the $SVM_{nlp}$ approach and the $SVM_{bow}$, respectively.

Table 6: Classification performances and the energy consumption results of different models on Unfair-ToS dataset.

| | Metric | $SVM_{bow}$ | $SVM_{nlp}$ | BERT | LegalBERT | Distil-BERT |
|---|---|---|---|---|---|---|
| | mF1 | 0.95 | 0.95 | 0.96 | 0.96 | 0.96 |
| | MF1 | 0.79 | 0.80 | 0.81 | 0.83 | 0.80 |
| Unfair-ToS | KWh | ×0.55 | 1 | ×62.11 | ×46.53 | ×30.05 |
| | € | ×0.55 | 1 | ×62.11 | ×46.53 | ×30.05 |
| | $CO_2$ | ×0.42 | 1 | ×21.85 | ×16.37 | ×10.57 |

## 4.2. The "in Production" Scenario

After the research and development phase (model selection by training-validation-test iterations) and the selection of the final solution, the model can be deployed in the production environment. Production is the final step of the machine learning life-cycle in industry. The model will be executed very frequently to analyze an actual stream of documents and produce predictions. In our analysis, we aimed to compare the energy requirements of different models when employed in the production step. For each model and each dataset, we performed the investigation for a standard bunch of documents. In particular, we considered a bunch composed of 100 documents as a representative of a real-world case. Accordingly, we evaluated the resource requirement in the prediction step by randomly selecting 100 documents from test splits of each dataset of the LexGLUE benchmark. Performance values (F1 scores) are not available in this analysis since they can only be evaluated in the R&D phase. The results are reported in Table 7.

In Table 7, we can see how the $SVM_{bow}$ approach reports the lowest values of energy consumption (and therefore cost and $CO_2$). However, the $SVM_{nlp}$ model is still an excellent solution showing energy indexes between 2-25 times the lightest $SVM_{bow}$. On the other hand, even in this phase, the BERT-based models have reported extremely high energy consumption values, in some cases even reaching a factor of 4000 times those of a standard $SVM_{bow}$.

Considering the excellent results obtained both in the R&D phase and in the "inproduction" scenario, the $SVM_{nlp}$ approach emerges as an excellent competitor, perfectly balancing performance (F1 scores very close to BERT-based models) and eco-friendly (energy consumption and optimal $CO_2$ emissions as the baseline $SVM_{bow}$ model).

Table 7: Comparison of time and energy consumption of the models for each dataset in the production scenario.

| | Metric | SVM$_{bow}$ | SVM$_{nlp}$ | BERT | LegalBERT | Distil-BERT |
|---|---|---|---|---|---|---|
| ECtHR A | Time | ∼**0.50sec** | ∼10sec | ∼20sec | ∼21sec | ∼13sec |
| | KWh | ×**1** | ×2.70 | ×577.06 | ×576.81 | ×342.32 |
| | €<br>CO | ×**1** | ×2.70 | ×577.06 | ×576.81 | ×342.32 |
| ECtHR B | Time | ∼**0.40sec** | ∼9sec | ∼21sec | ∼20sec | ∼13sec |
| | KWh | ×**1** | ×2.68 | ×665.10 | ×649.21 | ×394.62 |
| | €<br>CO | ×**1** | ×2.68 | ×665.10 | ×649.21 | ×394.62 |
| EUR-LEX | Time | ∼**0.1sec** | ∼2sec | ∼11sec | ∼11sec | ∼10sec |
| | KWh | ×**1** | ×2.14 | ×483.03 | ×533.24 | ×337.77 |
| | €<br>CO | ×**1** | ×2.14 | ×483.03 | ×533.24 | ×337.77 |
| LEDGAR | Time | ∼**0.02sec** | ∼1sec | ∼12sec | ∼12sec | ∼11sec |
| | KWh | ×**1** | ×5.11 | ×2523.67 | ×2640.76 | ×1743.25 |
| | €<br>CO | ×**1** | ×5.11 | ×2523.67 | ×2640.76 | ×1743.25 |
| SCOTUS | Time | ∼**1.4sec** | ∼1m20sec | ∼12sec | ∼13sec | ∼11sec |
| | KWh | ×**1** | ×4.40 | ×32.71 | ×34.72 | ×21.45 |
| | €<br>CO | ×**1** | ×4.40 | ×32.71 | ×34.72 | ×21.45 |
| Unfair ToS | Time | ∼**0.01sec** | ∼0.60sec | ∼11sec | ∼12sec | ∼10sec |
| | KWh | ×**1** | ×7.66 | ×3765.22 | ×4112.35 | ×3381.16 |
| | € | ×**1** | ×7.66 | ×3765.22 | ×4112.35 | ×3381.16 |
| | CO$_2$ | ×**1** | ×22.79 | ×503.67 | ×550.10 | ×452.29 |

## 5. CONCLUSIONS

In this paper, we present a comparative study of several models commonly used in text classification in terms of performance (F1), energy consumption (KWh), costs (€) and carbon footprint ($CO_2$), when used in a specific domain. In particular, we chose to address the "legal" area using the benchmark LexGLUE, a collection of seven datasets focused on the legal domain. For the analysis, we choose two families of models largely-used in the text classification task: LLM and SVM. From the first group (LLM-based models), we selected three BERT-based models: BERT, LegalBERT and DistilBERT; from the second group (SVM-based approaches), we chose a linear SVM which was investigated using two feature representations: the classic Bag-Of-Word (SVM$_{bow}$) and an advanced representation enriched with linguistic and semantic features (SVM$_{nlp}$).

The investigation aimed to explore the balance between performance and economic and ecological considerations of some different text categorization approaches when used in a real-world scenario. Thus, we carried out two types of investigations. We first considered an R&D scenario where a typical training/validation/test procedure is performed. Secondly, we considered the "in production" scenario, where the model has been deployed and is continuously invoked to analyze a real flow of documents and produce predictions.

The results show that a SVM$_{nlp}$ approach can achieve LLM performance in most of the LexGLUE datasets, with significant energy savings and $CO_2$ reduction. Elsewhere, concerns arise about how much additional energy can be justified for a few percent improvements in performance. Is this progress worth hundreds/thousands of energy consumption? Looking at the results obtained in both scenarios, the SVM$_{nlp}$ model seems to be an excellent ML solution, perfectly balancing performance (F1 scores very close to BERT-based models) and cost and eco-compatibility, with significant energy savings. Although some collaborative research on this subject has been presented in the literature [14], eco-friendly ML has yet to receive the attention it deserves. The trend towards larger deep neural networks should include also energy consumption and eco-friendly considerations, which should be significant points of this paradigm shift. The results presented in this work could lead machine learning researchers to include an environmental analysis in their activities.

Funding

The work was partially funded by:

- "enRichMyData - Enabling Data Enrichment Pipelines for AI-driven Business Products and Services", an Horizon Europe (HE) project, grant agreement ID: 101070284 [4];
- "IbridAI - Hybrid approaches to Natural Language Understanding", a project financed by the Regional Operational Program "FESR 2014-2020" of Emilia Romagna (Italy), resolution of the Regional Council n. 863/2021.

## REFERENCES

[1]     Stephan Bloehdorn and Andreas Hotho. Boosting for text classification with semantic features. In *Web Mining and Web Usage Analysis*, 2004.

[2]     Ilias Chalkidis, Ion Androutsopoulos, and Nikolaos Aletras. Neural legal judgment prediction in English. In *Proceedings of the 57th Annual Meeting of the Association for Computational Linguistics*, pages 4317–4323, Florence, Italy, July 2019. Association for Computational Linguistics.

[3]     Ilias Chalkidis, Manos Fergadiotis, and Ion Androutsopoulos. MultiEURLEX - a multi-lingual and multilabel legal document classification dataset for zero-shot cross-lingual transfer. In *Proceedings of the 2021 Conference on Empirical Methods in Natural Language Processing*, pages 6974–6996, Online and Punta Cana, Dominican Republic, November 2021. Association for Computational Linguistics.

[4]     Ilias Chalkidis, Manos Fergadiotis, Prodromos Malakasiotis, Nikolaos Aletras, and Ion Androutsopoulos. LEGAL-BERT: The muppets straight out of law school. In *Findings of the Association for Computational Linguistics: EMNLP 2020*, pages 2898–2904, Online, November 2020. Association for Computational Linguistics.

[5]     Ilias Chalkidis, Abhik Jana, Dirk Hartung, Michael Bommarito, Ion Androutsopoulos, Daniel Katz, and Nikolaos Aletras. LexGLUE: A benchmark dataset for legal language understanding in English. In *Proceedings of the 60th Annual Meeting of the Association for Computational Linguistics (Volume 1: Long Papers)*, pages 4310–4330, Dublin, Ireland, May 2022. Association for Computational Linguistics.

[6]     Corinna Cortes and Vladimir Vapnik. Support-vector networks. *Machine Learning*, 20(3):273–297, 1995.

[7]     Jacob Devlin, Ming-Wei Chang, Kenton Lee, and Kristina Toutanova. BERT: Pre-training of deep bidirectional transformers for language understanding. In *Proceedings of the 2019 Conference of the North American Chapter of the Association for Computational Linguistics: Human Language Technologies, Volume 1 (Long and Short Papers)*, pages 4171–4186, Minneapolis, Minnesota, June 2019. Association for Computational Linguistics.

[8]     Leonidas Gee, Andrea Zugarini, Leonardo Rigutini, and Paolo Torroni. Fast vocabulary transfer for language model compression. In *The 2022 Conference on Empirical Methods in Natural Language Processing*, Abu Dhabi, UAE, 12 2022.

[9]     Suyog Gupta, Ankur Agrawal, Kailash Gopalakrishnan, and Pritish Narayanan. Deep learning with limited numerical precision. In *International conference on machine learning*, pages 1737–1746. PMLR, 2015.

---

[10]  Marco Lippi, Przemysław Pałka, Giuseppe Contissa, Francesca Lagioia, Hans-Wolfgang Micklitz, Giovanni Sartor, and Paolo Torroni. Claudette: An automated detector of potentially unfair clauses in online terms of service. *Artificial Intelligence and Law*, 27(2):117–139, 2019.

[11]  Kadan Lottick, Silvia Susai, Sorelle A. Friedler, and Jonathan P. Wilson. Energy usage reports: Environmental awareness as part of algorithmic accountability. *CoRR*, abs/1911.08354, 2019.

[12]  Sasha Luccioni, Victor Schmidt, Alexandre Lacoste, and Thomas Dandres. Quantifying the carbon emissions of machine learning. In *NeurIPS 2019 Workshop on Tackling Climate Change with Machine Learning*, 2019.

[13]  Leonardo Rigutini. *Automatic Text Processing: Machine Learning Techniques*. LAP LAMBERT Academic Publishing, 07 2010. Saarbrücken, DE. isbn: 978-3-8383-7452-9 Book.

[14]  David Rolnick, Priya L. Donti, Lynn H. Kaack, Kelly Kochanski, Alexandre Lacoste, Kris Sankaran, Andrew Slavin Ross, Nikola Milojevic-Dupont, Natasha Jaques, Anna Waldman-Brown, Alexandra Luccioni, Tegan Maharaj, Evan D. Sherwin, S. Karthik Mukkavilli, Konrad P. Körding, Carla P. Gomes, Andrew Y. Ng, Demis Hassabis, John C. Platt, Felix Creutzig, Jennifer T. Chayes, and Yoshua Bengio. Tackling climate change with machine learning. *CoRR*, abs/1906.05433, 2019.

[15]  Victor Sanh, Lysandre Debut, Julien Chaumond, and Thomas Wolf. DistilBERT, a distilled version of BERT: smaller, faster, cheaper and lighter. *arXiv preprint arXiv:1910.01108*, 2019.

[16]  Victor Sanh, Lysandre Debut, Julien Chaumond, and Thomas Wolf. Distilbert, a distilled version of BERT: smaller, faster, cheaper and lighter. *CoRR*, abs/1910.01108, 2019.

[17]  Roy Schwartz, Jesse Dodge, Noah A. Smith, and Oren Etzioni. Green AI. *CoRR*, abs/1907.10597, 2019.

[18]  Sam Scott and Stan Matwin. Feature engineering for text classification. In *International Conference on Machine Learning*, 1999.

[19]  Fabrizio Sebastiani. Machine learning in automated text categorization. *ACM Comput. Surv.*, 34(1):1–47, mar 2002.

[20]  Harold J. Spaeth, Lee Epstein, Andrew D. Martin, Jeffrey A. Segal, Theodore J. Ruger, and Sara C. Benesh. 2020 supreme court database, version 2020 release 1.

[21]  Emma Strubell, Ananya Ganesh, and Andrew McCallum. Energy and policy considerations for deep learning in NLP. In *Proceedings of the 57th Annual Meeting of the Association for Computational Linguistics*, pages 3645–3650, Florence, Italy, July 2019. Association for Computational Linguistics.

[22]  Don Tuggener, Pius von Däniken, Thomas Peetz, and Mark Cieliebak. LEDGAR: A large-scale multi-label corpus for text classification of legal provisions in contracts. In *Proceedings of the 12th Language Resources and Evaluation Conference*, pages 1235–1241, Marseille, France, May 2020. European Language Resources Association.

[23]  Alex K. S. Wong, John W. T. Lee, and Daniel S. Yeung. Use of linguistic features in context-sensitive text classification. In *International Conference on Machine Learning and Computing*, 2005.

[24]  Michael Zhu and Suyog Gupta. To prune, or not to prune: exploring the efficacy of pruning for model compression. *arXiv preprint arXiv:1710.01878*, 2017.